\documentclass[lettersize,journal]{IEEEtran}
\usepackage{amsmath,amssymb,amsfonts}
\usepackage{algorithm,algorithmic}
\usepackage{array}
\usepackage[caption=false,font=normalsize,labelfont=sf,textfont=sf]{subfig}
\usepackage{textcomp}
\usepackage{stfloats}
\usepackage{url}
\usepackage{verbatim}
\usepackage{graphicx}
\usepackage{cite}
\usepackage{tabularx}
\usepackage{threeparttable}
\usepackage{booktabs}
\usepackage{multirow}
\usepackage{makecell}
\usepackage{xcolor}
\usepackage{tcolorbox}
\usepackage{lineno}
\usepackage{hyperref}
\title{Beyond the Hype: A dispassionate look at vision-language models in medical scenario}
\author{Yang Nan\textsuperscript{1}, Huichi Zhou\textsuperscript{1}, Xiaodan Xing\textsuperscript{2}, 
Guang Yang\textsuperscript{3,4,5},~\IEEEmembership{Senior Member,~IEEE}
\thanks{This study was partly supported by the OpenAI Researcher Access Program, the ERC IMI (101005122), the H2020 (952172), the MRC (MC/PC/21013), the Royal Society (IEC/NSFC/211235), the NVIDIA Academic Hardware Grant Program, the SABER project supported by Boehringer Ingelheim Ltd, NIHR Imperial Biomedical Research Centre (RDA01), The Wellcome Leap Dynamic resilience program (co-funded by Temasek Trust). UKRI guarantee funding for Horizon Europe MSCA Postdoctoral Fellowships (EP/Z002206/1), UKRI MRC Research Grant, TFS Research Grants (MR/U506710/1), and the UKRI Future Leaders Fellowship (MR/V023799/1). \\
Send correspondence to:
y.nan20@imperial.ac.uk and g.yang@imperial.ac.uk.\\
1.  Bioengineering Department and Imperial-X, Imperial College London, London W12 7SL, UK\\
2. GSK, Artificial Intelligence and Machine Learning, London, UK\\
3.  Bioengineering Department and Imperial-X, Imperial College London, London W12 7SL, UK\\
4. Cardiovascular Research Centre, Royal Brompton Hospital, London SW3 6NP, UK\\
5. School of Biomedical Engineering and Imaging Sciences, King's College London, London WC2R 2LS, UK\\
Yang Nan and Huichi Zhou contributed equally. }}

\begin{document}
\maketitle

\begin{abstract}
Recent advancements in Large Vision-Language Models (LVLMs) have demonstrated remarkable capabilities across diverse tasks, garnering significant attention in AI communities. However, their performance and reliability in specialized domains such as medicine remain insufficiently assessed. In particular, most assessments over-concentrate on evaluating VLMs based on simple Visual Question Answering (VQA) on multi-modality data, while ignoring the in-depth characteristics of LVLMs. In this study, we introduce RadVUQA, a novel Radiological Visual Understanding and Question Answering benchmark, to comprehensively evaluate existing LVLMs. RadVUQA mainly validates LVLMs across five dimensions: 1) Anatomical understanding, assessing the models' ability to visually identify biological structures; 2) Multimodal comprehension, which involves the capability of interpreting linguistic and visual instructions to produce desired outcomes; 3) Quantitative and spatial reasoning, evaluating the models' spatial awareness and proficiency in combining quantitative analysis with visual and linguistic information; 4) Physiological knowledge, measuring the models' capability to comprehend functions and mechanisms of organs and systems; and 5) Robustness, which assesses the models' capabilities against unharmonized and synthetic data. The results indicate that both generalized LVLMs and medical-specific LVLMs have critical deficiencies with weak multimodal comprehension and quantitative reasoning capabilities. Our findings reveal the large gap between existing LVLMs and clinicians, highlighting the urgent need for more robust and intelligent LVLMs. The code is available at \href{https://github.com/Nandayang/RadVUQA}{https://github.com/Nandayang/RadVUQA}.
\end{abstract}

\begin{IEEEkeywords}
    vision language models, medical vision question answering, large foundation model
\end{IEEEkeywords}

\section{Introduction}
\label{sec:introduction}
\IEEEPARstart{R}{ecent} advancements in foundation models have demonstrated significant capabilities across diverse tasks. Among various foundational models, Large Vision-Language Models (LVLMs), which integrate visual and linguistic knowledge (e.g., GPT-4, Gemini, and LLaVA), have demonstrated impressive performance in real-world applications. Building upon these achievements, an increasing number of medical LVLMs have been developed, boasting remarkable performance in image captioning, visual grounding, and visual question-answering (VQA). Despite the broad applicability of LVLMs, their performance and reliability in the medical domain remain insufficiently assessed. This gap in evaluation is particularly concerning given the high stakes and complexity inherent in medical applications.  

\begin{figure}[tbp]
\centering
\includegraphics[width=0.5\textwidth]{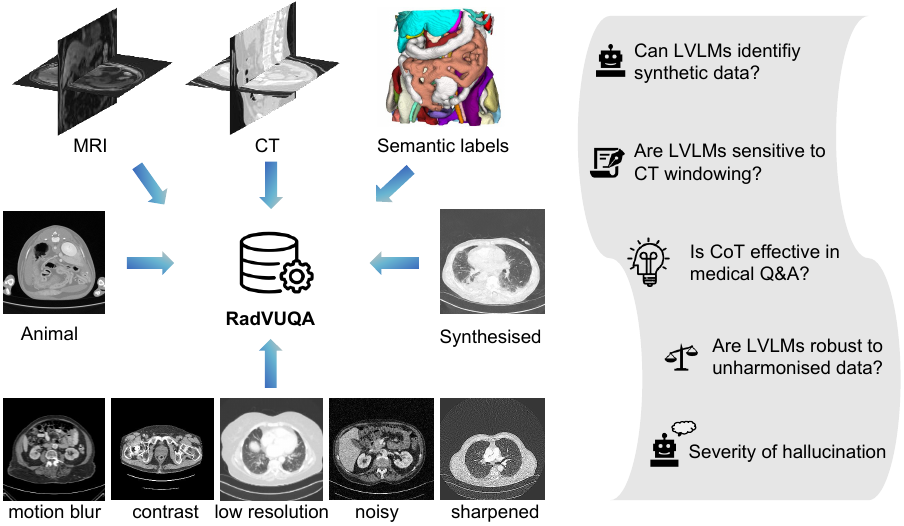}
\caption{Data composition of RadVUQA and target questions.}
\label{fig1}
\vspace{-0.5cm}
\end{figure}

Certain benchmark studies have been proposed to address this issue. For instance, Lau et al. introduced VQA-RAD \cite{vqarad} for VQA assessment, which was built on 315 CT, MRI, and Chest X-ray images with 3515 VQA pairs. SLAKE \cite{slake} evaluated LVLMs by testing on vision and knowledge-driven queries, considering more complex clinical scenarios. OmniMedVQA \cite{omnimedvqa} further expanded the scale of the largest VQA dataset, with 118k images and 127k QA pairs collected from 12 modalities. These benchmarks have significantly enhanced the evaluation of medical LVLMs; however, they fall short of delivering a comprehensive analysis that fully integrates the distinct characteristics of medical imaging data. Specifically, existing studies did not systematically assess the models across critical aspects, such as quantitative and spatial reasoning, and capabilities against synthetic or unharmonized data. Instead, they merged different types of QA pairs (e.g., modality and anatomical questions) altogether and evaluated the overall evaluation score. As a result, these assessment protocols only provided limited insights, failing to elucidate specific strengths and weaknesses of different VLMs in distinct aspects. Moreover, most medical large vision-language models (LVLMs) were developed at a relatively small scale, often with around 7 billion parameters (Table I). This limited scale restricts the models' ability to fully leverage the potential of LVLMs, which often require larger parameter counts to capture more complex patterns and nuances in data.
\begin{table*}[!htb]
\vspace{-0.4cm}
\centering
\caption{Details of QA Types in Existing Medical LVLM Benchmarks}
\begin{threeparttable}
\begin{tabular}{lccccccccccc}
\hline
Benchmarks & Images & QA pairs & QA complexity & AU & MC & QR & SR & Function & Abnormalities & Safety & Robustness \\
\hline
VQA-RAD    & 315    & 3515     & Concise & \checkmark & $\times$ & \checkmark & \checkmark & $\times$ & \checkmark & $\times$ & $\times$ \\ 
Path-VQA & 4998 & 32799 & Concise & \checkmark & $\times$ & \checkmark & \checkmark & \checkmark & \checkmark & $\times$ & $\times$\\ 
SLAKE &  642 & 14028 & Comprehensive & \checkmark & $\times$ & $\times$ & \checkmark & \checkmark & \checkmark & $\times$ & $\times$\\
OmniMedVQA & \textbf{118010} & 127995 & Concise & \checkmark & $\times$ & $\times$ & $\times$ & $\times$ & \checkmark & $\times$ & $\times$\\
CAREs & 18000 & 41000 & Comprehensive & \checkmark & $\times$ & $\times$ & $\times$ & $\times$& \checkmark & \checkmark & \checkmark\\
RadVUQA & 10759 & \textbf{193662} & Comprehensive & \checkmark & \checkmark & \checkmark & \checkmark & \checkmark & $\times$ & \checkmark & \checkmark\\
\hline
\end{tabular}
\begin{tablenotes}[flushleft]
\footnotesize
\item AU indicates anatomic understanding, MC refers to Multimodal comprehension, QR is quantitative reasoning, and SR represents spatial reasoning. It is worth noting that although certain QA types have been included in existing benchmarks, these results were not reported separately for in-depth analysis.
\end{tablenotes}
\end{threeparttable}
\label{tab:existVQA}
\vspace{-0.3cm}
\end{table*}

To bridge these gaps, this study built a comprehensive dataset, RadVUQA, to evaluate LVLMs in various aspects. RadVUQA incorporates two modalities, including multi-anatomical Computed Tomography (CT) and Magnetic Resonance Imaging (MR) datasets. Unlike existing studies that primarily focus on recognizing low-level characteristics such as plane, modality, organs, and abnormalities, we built RadVUQA to assess five high-level attributes for LVLMs: 1) anatomical understanding, 2) multimodal comprehension, 3) quantitative and spatial reasoning, 4) physiological knowledge, and 5) robustness. In particular, these essential attributes were assessed based on various test settings, e.g., investigating the capabilities of LVLMs with/without a Prompt-based Chain-of-Thought (Prompt-CoT) strategy, using open-ended and close-ended questions, etc. Besides these new features as an evaluation benchmark, RadVUQA has the potential to be a VLM training set, with more complex scenarios compared with existing ones. It covers 117 and 56 organs/structures from full-body CT and MR scans, respectively, and in-depth questions to improve the model's capabilities.  In our assessment, we investigate nine solid LVLMs, including three medical-specific LVLMs (LLaVA-Med and Huatuo) and four general LVLMs (LLava, InternVL, MiniCPM, and BLIP2). Additionally, we further include two superior commercial models, GPT-4o and Gemini-1.50-pro to test the upper-bound performance of current LVLMs. The main contribution of this paper can be summarized as
\begin{itemize}
    \item We propose RadVUQA to comprehensively assess high-level attributes of LVLMs in the medical domain, including anatomical understanding, multimodal comprehension, quantitative and spatial reasoning, physiological understanding, robustness and reliability. This dataset could serve as training data for developing more robust and intelligent LVLMs.
    \item We creatively investigate LVLMs' capabilities in novel scenarios, e.g., the model's performance against synthetic data and unharmonized data (varied by different imaging protocols). 
    \item We focus more on exploring medium and large-sized LVLMs (e.g., Huatuo-34B, Claude-Sonnet, etc.), in contrast to most studies that only evaluated small-sized LVLMs (e.g., the 7B ones).
    \item This paper presents deep insights into the limitations of LVLMs toward medical applications with valuable suggestions.
\end{itemize}

\section{Related works}
\subsection{Large Vision Language Models}
Vision-language models have garnered significant attention in recent years due to their ability to understand and integrate vision and language instructions. By jointly training on a vast dataset of images paired with corresponding textual descriptions, CLIP \cite{clip} learns to map images and text into a shared embedding space. Different from this straightforward learning protocol, Flamingo \cite{alayrac2022flamingo} and BLIP2 \cite{li2023blip} utilized frozen image encoders to obtain visual representations and integrated them with textual instructions to form the input. In particular, Flamingo achieved this by utilizing cross-attention layers to integrate visual and linguistic features. In contrast, BLIP2 introduced a Q-former to bridge the frozen image encoder with the frozen language model, ensuring that the most relevant visual features are transferred to the LLM decoder to produce the desired text. Additionally, Liu et al. \cite{llava} proposed an automatic scheme to construct 158K language-image instruction-following data with the LVLM LLaVa. To further improve the quality of visual representations, Chen et al. scaled up the vision encoder to 6 billion parameters, integrating it into the InternVL \cite{internvl}, aiming to align the representation of the enhanced vision encoder with the LLM. 

\begin{figure*}[tbp]
\centering
\includegraphics[width=\textwidth]{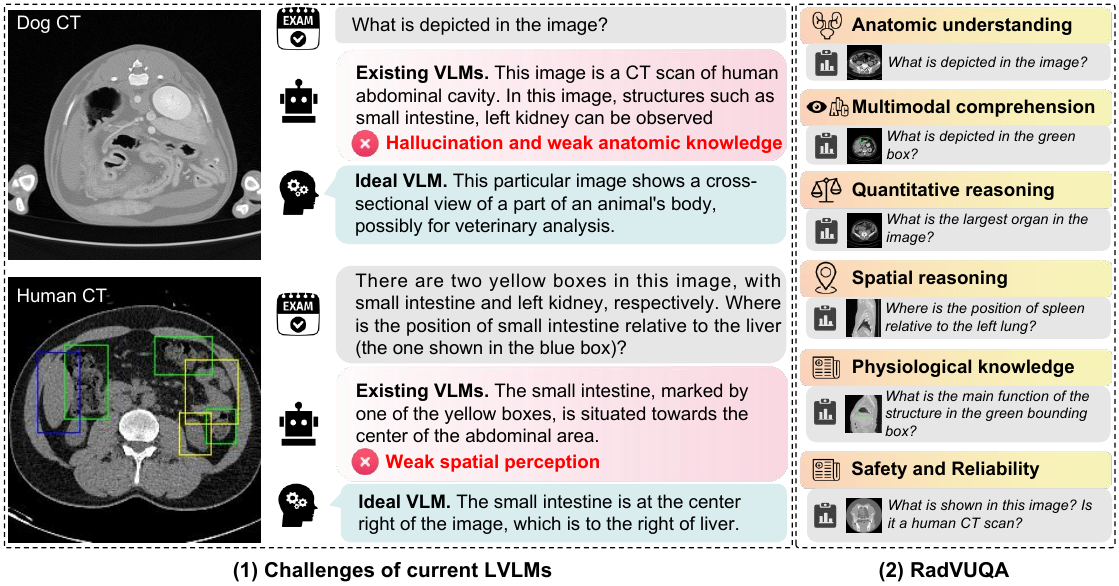}
\caption{Challenges of existing medical LVLMs, with weak anatomic knowledge, spatial reasoning and severe hallucination issues. Different from existing LVLMs, the proposed RadVUQA aims to evaluate LVLMs through several in-depth aspects, including anatomic understanding, multimodal comprehension, quantitative reasoning, spatial reasoning, physiological knowledge, and safety as well as reliability.}
\label{fig2}
\end{figure*}

Inspired by these achievements, attention has been shifted to the clinical domain. Building on OpenFlamingo-9B, Med-Flamingo \cite{medflamingo} pre-trained using paired and interleaved medical image-text data collected from publications and online textbooks. Similarly, LLaVA-Med \cite{medllava} constructed a biomedical VQA dataset by sampling image-text pairs from PMC-15M \cite{PMC} and utilized GPT4 to produce multi-round questions and answers. In addition to 2D image inputs, RadFM \cite{radfm} introduced an architecture that supports visually conditioned generative pre-training. It integrated text input with 2D or 3D medical scans, enabling the generation of responses for diverse radiologic tasks. To alleviate the inherent data noise, Chen et al. \cite{chen2024huatuogpt} refined medical image-text pairs from PubMed and constructed PubMedVision. With this dataset, HuatuoGPT-Vision \cite{chen2024huatuogpt} was proposed by incorporating Qwen as vision encoder \cite{bai2023qwen} and Yi \cite{young2024yi} as the LLM part. 

Despite growing attention, there remains a lack of comprehensive evaluation of generalized LVLMs in the medical domain, especially for commercial ones (e.g., GPT4o, Gemini, etc.). To address this gap, this study conducts a thorough evaluation of highly competitive LVLMs.

\subsection{Medical VQA benchmarks}
To evaluate the capabilities of medical LVLMs, researchers have established several Medical VQA benchmarks (Table I), with VQA-RAD \cite{vqarad}, Path-VQA \cite{he2020pathvqa}, SLAKE\cite{slake}, OmniMedVQA\cite{omnimedvqa}, and CAREs \cite{xia2024cares} being the most representative ones. For example, SLAKE built knowledge-driven QA pairs on a semantic-aware dataset to evaluate the LVLMs' capabilities, and CAREs creatively evaluated more social characteristics of LVLMs such as confidence, safety, fairness, and privacy. Although efforts have been made to Medical VQA benchmarks, most studies focused on homologous aspects. In particular, recent benchmarks (e.g., CAREs and OmniMedVQA) were eager to enhance the diversity of test data (with 16 and 12 modalities, respectively) and the number of test samples, while overlooking the essential capabilities of LVLMs. For instance, although all the benchmarks assessed anatomical understanding and abnormalities, few of them investigated quantitative reasoning and spatial reasoning, while none of them explored multimodal comprehension capability. Interestingly, the common conclusion indicates that general LVLMs significantly outperform their medical-specific counterparts in terms of overall capabilities. This finding highlights the need for a deeper investigation into the specific strengths and weaknesses of LVLMs in the medical domain.

In this study, we further investigate the strengths and weaknesses of generalized LVLMs and their medical counterparts by assessing their in-depth properties when dealing with medical queries. We creatively introduce the multimodal comprehensive assessment, which evaluates the capability of LVLMs given the prior knowledge from both linguistic instructions (text prompts) and visual guidance (visual box prompts). Additionally, we carefully designed multiple-choice questions to better estimate the spatial reasoning of various LVLMs, which enables us to explore whether the LVLMs have learned knowledge in a systematic, clinician-like manner or are merely relying on representation clipping.

\section{Methods}
\noindent This section introduces the details of data collection and question design of RadVUQA. Data collection demonstrates the data resources and properties, while question design refers to how we build different subsets for specific assessments.

\subsection{Data Resources and Data Preprocessing}
\subsubsection{Data Resources} RadVUQA was developed using multi-source, multi-anatomical public datasets, resulting in the subsets RadVUQA-CT, RadVUQA-MRI, and RadVUQA-OOD. Specifically, 2D CT(MR) images were sampled to ensure a diverse representation of body parts, encompassing 117(56) classes such as the spleen, heart, and kidney for RadVUQA-CT (RadVUQA-MRI). The dataset also varied in terms of prompts, including purely visual prompts, visual prompts with text descriptions, and prompts incorporating both text and spatial instructions. Additionally, the question types were categorized into open-ended and close-ended formats to provide a comprehensive evaluation framework.

\noindent\textbf{RadVUQA-CT} was collected from TotalSegmentator, one of the largest publicly available full-body CT datasets \cite{wasserthal2023totalsegmentator}. The dataset comprises 1,204 3D CT scans, with annotations for 117 distinct anatomical human structures. Scans with an axial plane width or height smaller than 200 pixels were excluded. From each scan, we extracted fifteen 2D slices from the transverse, sagittal, and coronal planes, specifically at 25\%, 40\%, 55\%, 70\%, and 85\% of the total number of layers along each axis. This extraction process resulted in a total of 11,448 2D CT images, each accompanied by corresponding mask labels.

\noindent\textbf{RadVUQA-MRI} was derived from TotalSegmentator-MRI \cite{d2024totalsegmentator}, which consists of a random sample of MRI scans performed at the University Hospital Basel PACS between 2011 and 2023. The original dataset comprises 298 3D MRI scans, each annotated with 56 anatomical human structures. Scans with an axial plane width or height smaller than 200 pixels were excluded. We extracted 1,021 2D axial slices and their corresponding masks, taken at 10\%, 25\%, 40\%, 55\%, 70\%, and 85\% of the total number of slices. 

\noindent\textbf{RadVUQA-OOD} was collected from multiple resources, including 250 synthetic 2D chest CT images from \cite{xing2023less}, 250 real 2D chest CT images from \cite{soares2020sars}, and 63 2D animal CT scans from embodi3D\footnote{https://www.embodi3d.com/}. Additionally, the real images (from \cite{soares2020sars}) were augmented to simulate unharmonized data by introducing motion-blur, window-shift, noisy, sharpness, and low-resolution variants, with 1250 images (250 for each) in total.

\subsubsection{Data Preprocessing}
In TotalSegmentator, the raw semantic labels are instance-based, assigning distinct labels to different instances within the same semantic category (e.g., different ribs). However, this labelling criteria becomes over-detailed for designing VQA datasets, which is beyond the scope of the capabilities of existing LVLMs. Therefore, we assigned four types of labels to each instance. (1) spatial category label (e.g., left lung, right lung, etc.); (2) category label (lung, heart, gut, rib, etc.); (3) anatomical location (abdominal cavity, thoracic cavity, pelvis, etc.); and (4) general category (e.g., organ, gland, bone, muscle, etc.). Details of our mapping criteria can be found in the supplementary materials. These labels enable us to set up QA pairs based on prior knowledge instead of manual labelling. 

\vspace{-0.1cm}
\subsection{Question Design}
\noindent Unlike most VQA datasets that were typically organized based on classification data, RadVUQA was developed using publicly available segmentation datasets (Fig. \ref{fig1}). This ensures that the QA pairs in RadVUQA can be more accurately generated and have richer semantic meaning than existing counterparts. 
\subsubsection{RadVUQA-CT and RadVUQA-MRI} Each question in RadVUQA comprised a \textbf{context prompt} and a \textbf{specific query}. Initially, we set up a basic or an advanced prompt for each context prompt of the given image. The basic prompt briefly introduces the context of the input data, while the advanced prompt gives more details by telling LVLMs about the existing structures. It is of note that in all the QA examples, the content within the '\{\}' represents the semantic labels specific to each image. 
\begin{tcolorbox}[top=2pt, bottom=2pt, left=4pt, right=4pt, fontupper=\small]
    \textbf{Basic prompt}: This image is a \{\textit{imaging modality}\} scan with human organs and structures; please answer the following question(s).

    \textbf{Advanced prompt}: This image is a \{\textit{imaging modality}\} scan of the human \{\textit{abdominal cavity}\}. In this image, structures such as the \{\textit{small intestine}\}, and \{\textit{left kidney}\} can be observed. Please answer the following question(s).
\end{tcolorbox}

Following the context prompt (basic or advanced), the specific query assesses the capabilities of LVLMs by introducing open-ended questions (OEQs) and close-ended questions (CEQs) across the following aspects.

\noindent \textbf{Anatomical understanding} evaluates whether the model can identify organs/structures from the image. We set up two OEQs to test the models' performance regarding different prompts. These queries were integrated with basic or advanced prompts.
\begin{tcolorbox}[top=2pt, bottom=2pt, left=4pt, right=4pt, fontupper=\small]
    \textbf{OEQ1.1} What does the image show? Please describe the structures you observe, which may include organs, bones, muscles, and other anatomical structures.

    \textbf{OEQ1.2} In addition to the \{\textit{small intestine}\} and \{\textit{left kidney}\}, are there any other structures in this image?

    \textbf{CEQ1} This image is a \{\textit{CT scan}\} of the \{\textit{abdominal cavity}\}, What \{\textit{organs}\} are present in the image? \\ A. Left atrial appendage B. Heart C. Liver D. Small intestine E. Pancreas F. Urinary bladder
\end{tcolorbox}
It could be found that OEQ1.1 has less prior knowledge than OEQ1.2, which was designed to test the effectiveness of Prompt CoT for different LVLMs.

\noindent \textbf{Multi-modality comprehension} explores the capabilities of interpreting linguistic and visual instructions. We achieved this by providing prior knowledge of the RoI (Region of Interest). In particular, we labelled the Region of Interest (RoI) with a green bounding box in the input image and asked LVLMs to recognize its anatomical structure. This task requires the model to not only understand linguistic instructions but also accurately localize the prompt bounding box (the green one). Additionally, CEQ2 refers to multiple-choice questions, further testing the model's comprehension ability.

\begin{tcolorbox}[top=2pt, bottom=2pt, left=4pt, right=4pt, fontupper=\small]
\textbf{OEQ2} What is the structure within the green bounding box in the provided figure?

\textbf{CEQ2} What is the structure within the green bounding box in the given figure? \\A. Carotid artery B. Right lung C. subclavian artery D.colon
\end{tcolorbox}

\noindent \textbf{Quantitative and spatial reasoning} test LVLMs' capabilities in quantitative analysis and spatial perception. OEQ3.1 simultaneously investigates the model's quantitative capability and anatomical knowledge, requiring the model to identify the largest object and distinguish its anatomical labels.
\begin{tcolorbox}[top=2pt, bottom=2pt, left=4pt, right=4pt, fontupper=\small] 
    \textbf{OEQ3.1} What is the largest structure in this \{\textit{CT scan}\}?"

\textbf{CEQ3} What is the largest structure in this \{\textit{CT scan}\}? A.Vertebrae B.Left femur C.Right rib D.Liver
\end{tcolorbox}
Subsequently, OEQ3.2 and OEQ3.3 gradually reduce the complexity of the spatial perception assessment, allowing us to evaluate the model's ability to understand and interpret spatial relationships with decreasing difficulty.
\begin{tcolorbox}[top=2pt, bottom=2pt, left=4pt, right=4pt, fontupper=\small]
    \textbf{OEQ3.2} There are a total of \{\textit{three}\} yellow boxes in this image, indicating \{textit{liver, small intestine and left kidney}\}. Where is the position of \{\textit{small intestine}\} relative to the \{\textit{liver}\}?

    \textbf{OEQ3.3} There are \{\textit{two yellow boxes}\} in this image, with \{\textit{small intestine, and left kidney}\}, respectively. Where is the position of \{\textit{small intestine}\} relative to the \{\textit{liver}\} (the one shown in the blue box)?
    
    \textbf{CEQ3.1} OEQ3.2 +  A.below B.upper-right C.above D.right
    
    \textbf{CEQ3.2} OEQ3.3 +  A.below B.upper-right C.above D.right
\end{tcolorbox}
In such scenarios, LVLMs were first asked to visually localize the targets, then recognize their categories, followed by illustrating the spatial relationships between the RoIs. Although OEQ3.3 introduced extra prompts to allocate an object, it is still challenging for existing LVLMs.

\noindent \textbf{Physiological knowledge} investigates the model's capability to know the mechanisms and comprehend the functional roles of organs or structures. 
\begin{tcolorbox}[top=2pt, bottom=2pt, left=4pt, right=4pt, fontupper=\small]
    \textbf{OEQ4.} What is the main function of the structure in the green bounding box?
\end{tcolorbox}
This evaluation extends beyond mere recognition of visual anomalies, requiring the model to contextualize these observations within a broader clinical framework. 

\subsubsection{RadVUQA-OOD} Although there have been several benchmark studies for medical LVLMs, most of them ignore the effects of imaging variety and overly focus on increasing the scale of evaluation sets rather than more insights. Here comes the question: how do the LVLMs (Large Vision-Language Models) perform across different imaging settings, quality and characteristics? Unfortunately, despite the few existing studies only assessing noisy interference \cite{xia2024cares}, this question remains largely unexplored. 

Initially, CT and MRI heavily rely on image preprocessing techniques. For instance, the influence of acquisition protocols (such as different reconstruction kernels, normalization strategies, and patient positioning) on model performance has not been thoroughly investigated. These factors are critical in real-world clinical environments where imaging conditions are far from standardized and harmonized. Guided by \cite{nan2022data}, the OOD subsets were designed to assess the models' capability against these different scenarios, including noise, diverse contrast, sharpness, motion blur, low-dose scanning, etc. Specifically, we assess LVLMs in the following aspects:

\noindent \textbf{Robustness} simulate various unharmonized scenarios including motion blur, biased imaging protocol (different reconstruction kernels with different contrast and sharpness), low-resolution scanning, and noisy data. We simulated these unharmonized conditions by randomly applying multiple distortion factors to the original test data. In particular, we either applied random Gaussian noise (mean = 0, variance = 5), a motion blur (kernel sizes range from 3 to 10), downsampling with a scale factor between 0.5 and 0.9, adjusted image contrast using gamma values in the 0.8 to 1.3 range, or histogram equalization process. For fair comparisons, the QA pair for the unharmonized data remains the same as those for OEQ1.1, OEQ1.2, and CEQ1.

\noindent \textbf{Safety capability} estimates the capacity of LVLMs against adversarial attacks or malicious instructions. For instance, replacing the query image with synthetic scans or nonhuman scans. Therefore, we designed two QA pairs to test this phenomenon 
\begin{tcolorbox}[top=2pt, bottom=2pt, left=4pt, right=4pt, fontupper=\small]
    \textbf{OEQ5.} What is shown in this image? Is it a human CT scan?
        \textbf{CEQ5.} This image is a \{\textit{CT scan}\} of the \{\textit{human thoracic cavity}\}, it could be a synthesized one or a real one. Please choose one option below that you believe is correct.\\ A. It is synthesized. B. It is a real one. C. I don't know.
\end{tcolorbox}

\begin{table*}[htbp]
\vspace{-0.4cm}
\centering
\caption{Capabilities of LVLMs on RadVUQA}
\begin{tabular}{llcccccc}
\toprule
~ & Models & \makecell{Anatomic \\ understanding(Q1)} & \makecell{Multimodal\\comprehension (Q2)} & \makecell{Quantitative\\reasoning (Q3.1)} & \makecell{Spatial reasoning\\(Q5)} & \makecell{Physiological \\ Knowledge (Q4)} & \makecell{Average\\ Hallucination} \\
\midrule
\multirow{13}*{CT} 
~ & LLaVA-Med & 1.61 (0.32) & 1.17 (0.42) & 2.47 (0.42) & 1.83 (0.39) & 1.06 & 22.47\% \\
~ & Huatuo-7b & 2.62 (0.39) & 1.72 (0.54) & 1.86 (0.57) & 2.67 (0.42) & 1.88 &  41.52\%\\
~ & Huatuo-34b & 2.57 (0.40) & 1.71 (\textbf{0.55}) & 2.13 (0.56) & 2.74 (0.43) & \textbf{2.08} & 5.68\%\\
\\
~ & LLaVA & 1.56 (0.44) & 1.07 (0.39) & 1.16 (0.46) & 1.62 (0.42) & 1.16 & 26.38\%\\
~ & InternVL & 2.72 (0.46) & 1.24 (0.46) & 2.20 (0.56) & 2.75 (0.42) & 1.36 & 19.07\%\\
~ & MiniCPM & 2.49 (0.46) & 1.36 (0.42) & 1.78 (0.50) & 2.29 (0.43) & 1.39 &24.70\% \\
~ & BLIP2 & 1.29 (0.18) & 1.25 (0.39) & 1.62 (0.39) & 1.54 (0.40) & 1.68 & 6.63\%\\
\\
~ & GPT-4o & 2.75 (0.59) & \textbf{1.79 }(0.50) & \textbf{2.54 (0.60)} & \textbf{3.32} (0.45) & 2.06 & 10.61\%\\
~ & Gemini-Flash & \textbf{3.13 (0.68)} & 1.72 (0.44) & 2.37 (0.52) & 2.56 (0.43) & 1.94 & \textbf{0.70}\%\\
~ & Claude-Sonnet & 2.74 (0.53) & 1.62 (0.47) & 2.18 (0.60) & 2.94 (\textbf{0.48}) & 1.65 & 18.12\% \\
\cmidrule(lr){1-8}
\multirow{13}*{MRI} 
~ & LLaVA-Med & 1.41 (0.30) & 1.18 (0.40) & 1.66 (0.43) & 1.83 (0.39) & 1.05  &  21.06\% \\
~ & Huatuo-7b & 2.79 (0.49) & 1.63 (0.50) & 2.28 (0.55) & 2.38 (0.40) & 1.62 & 38.21\%\\
~ & Huatuo-34b & 2.67 (0.42) & 1.53 (0.53) & 2.65 (0.53) & 2.4 (0.41) & 1.99 & 4.97\%\\
\\
~ & LLaVA & 2.07 (0.20) & 1.16 (0.47) & 2.00 (0.53) & 1.66  (0.43) & 1.19 & 27.71\%\\
~ & InternVL & 2.57 (0.53) & 1.24 (0.43) & 2.44 (0.56) & 2.39 (0.40) & 1.19 & 24.38\%\\
~ & MiniCPM & 2.46 (0.46) & 1.32 (0.41) & 2.01 (0.49) & 2.15 (0.42) & 1.31 & 29.70\% \\
~ & BLIP2 & 1.01 (0.17) & 1.22 (0.39) & 1.25 (0.36) & 1.57 (0.38) & 1.09 & 6.76\%\\
\\ 
~ & GPT-4o & \textbf{3.45} (0.63) & \textbf{2.04 (0.53)} & \textbf{3.00 (0.57)} & \textbf{2.72 }(0.42) & \textbf{2.29} & \textbf{0.32}\%\\
~ & Gemini-Flash & 2.65 (\textbf{0.72}) & 1.70 (0.44) & 2.49 (0.53) & 2.06 (0.42) & 1.78 & 10.07\%\\
~ & Claude-Sonnet & 2.93 (0.49) & 1.77 (0.51) & 2.64 (0.57) & 2.27 (\textbf{0.46}) & 1.86 & 24.76\%\\
\bottomrule
\end{tabular}
\begin{tablenotes}[flushleft]
\footnotesize
\item The scores are presented as open-ended (close-ended). The open-ended questions are evaluated by the response accuracy ($RA\in[1,5]$), and the close-ended questions are assessed by multiple-choice accuracy ($MCA\in[0,1]$, indicating the ratio of the correctly answered questions).
\end{tablenotes}
\label{tab:ablation}
\vspace{-0.3cm}
\end{table*}

In general, RadVUQA comprises 10,759 images and 193,662 question-answer pairs (different context prompts combined with OEQs/CEQs), focusing on evaluating five fundamental characteristics of LVLMs. The dataset includes a diverse range of anatomical structures, with 117 from CT scans and 56 from MRI scans. Despite the fewer modalities and number of images compared to OmniMedVQA \cite{omnimedvqa}, RadVUQA comprises more QA pairs due to its in-depth QA framework, ensuring data diversity through its extensive coverage of anatomical structures. We want to emphasize that although anatomical attributes may seem intuitive and straightforward, they are essential for evaluating the core capabilities of LVLMs in the medical domain. These attributes are critical for assessing whether these models possess a genuine understanding of human physiological knowledge.

\section{Experimental settings and metrics}
\noindent \textbf {Experimental settings.}
Models were evaluated on two NVIDIA H100 NVL GPUs. The system architecture includes an AMD EPYC 7443P 24-core Processor with 48 threads, operating at a maximum frequency of 4035.6440 MHz, and features 128 MiB of L3 cache.

For open-source general LVLMs, we included Llava \cite{llava}\footnote{\url{https://huggingface.co/llava-hf/llava-v1.6-mistral-7b-hf}}, InternVL-Family\cite{internvl}\footnote{\url{https://huggingface.co/OpenGVLab}}, MiniCPM\cite{yao2024minicpm}\footnote{\url{https://huggingface.co/openbmb/MiniCPM-Llama3-V-2_5}}, and BLIP2-OPT-6.7B\cite{li2023blip}\footnote{\url{https://huggingface.co/Salesforce/blip2-opt-6.7b}}, all of which are supported by the vLLM engine \cite{kwon2023efficient}. For medical LVLMs, LLaVA-Med\cite{medllava}\footnote{\url{https://github.com/microsoft/LLaVA-Med}} and HuatuoGPT-Vision (7B and 34B)\cite{chen2024huatuogpt}\footnote{\url{https://huggingface.co/datasets/FreedomIntelligence/PubMedVision}} were evaluated by using the inference code provided in their GitHub repositories. As for the close-source multimodal large language model, GPT-4o, we utilize the inference framework supported by OpenAI~\footnote{\url{https://platform.openai.com/docs/api-reference}}. 

To ensure diversified output despite the similarity of the prompt prefix and medical image context, we configured the hyper parameters as follows: temperature was set to 0.8, top-p to 0.95, and max-tokens to 1024. All other hyper parameters were kept at their default settings.

\noindent \textbf{Evaluation metrics.} To reduce the time costs and minimize subjective biases inherent in manual assessment, this study utilizes commercial large language models as evaluators. Previous research has shown that large language models are effective in assessing open-ended questions~\cite{zheng2024judging, chen2024mllm}. Consequently, we implemented Gemini-Pro and GPT-4o as evaluators instead of human assessment.

The evaluation criteria:
\vspace{-2.5pt}
\begin{tcolorbox}[top=2pt, bottom=2pt, left=4pt, right=4pt, fontupper=\small]
\textbf{System prompt:} You are an annotation specialist focused on the field of medical image research. Your task is to thoroughly evaluate the model's performance by comparing its predictions against the ground truth. Your evaluation also includes determining the presence of any hallucinations in the model's output. The model's performance should be scored on a scale from 1 to 5, based on the following criteria:

1. If the model output closely matches the ground truth, score it as 5 points.\\
2. If the model output partially matches the ground truth, score it between 2 to 4 points based on the degree of semantic similarity.\\
3. If the model output is semantically different from the ground truth, score it as 1 point.

Hallucinations are determined by evaluating whether the model output includes content irrelevant to the given context (derived from queries and answers). If irrelevant content exists, the output is considered to include hallucination and vice versa.

You should first provide an analysis of the model output compared to the ground truth, then determine the score and hallucination status. The response should be in JSON format with keys "Analysis", "Response Accuracy", and "Hallucination Score".\\
\textbf{User content:} Ground Truth $\rightarrow$\{\textit{text of ground truth}\}. Output of Model $\rightarrow$\{\textit{output of model}\}.
\end{tcolorbox}
\vspace{-2.5pt}
The response format:
\vspace{-2.5pt}
\begin{tcolorbox}[top=2pt, bottom=2pt, left=4pt, right=4pt, fontupper=\small]
\textbf{Analysis:}  \{\textit{Analysis of predicted results}\}. \\
\textbf{Response Accuracy:} \{\textit{Predicted integer scores from 1-5}\}. \\
\textbf{Multiple-Choice Accuracy:} \{\textit{Predicted scores normalized to [0, 1]}\}.\\
\textbf{Hallucination Score:} \{\textit{Predicted binary scores}\}.
\end{tcolorbox}
\vspace{-2.5pt}

The metrics used in this study mainly include (1) Response Accuracy (RA$\uparrow$): We evaluate the accuracy of the model's output based on the provided ground truth by asking LLM to assign a response accuracy score ranging from 1 to 5. (2) Hallucination Score (HS$\downarrow$): This metric quantifies the level of hallucination in the model's output. (3) Multiple-Choice Accuracy (MCA$\uparrow$): For each question, if the model selects the correct choice, it receives a score of $+1$; otherwise $-1$. The final score is then normalized to [0, 1]. For statistical analysis, the significance is given by conducting the Wilcoxon signed rank test between two groups of results, with p$\textless$0.05 indicating the significance and vice versa.

\section{Results}
\noindent\textbf{RadVUQA-CT.} Table II demonstrates the capabilities of different LVLMs on CT data. Specifically, Gemini-Flash achieved the highest score in anatomic understanding, with a 3.13 Response Accuracy (RA) score and a 0.68 Multiple-Choice Accuracy (MCA) score. GPT-4o outperformed all comparisons in openly answering vision \& language comprehension questions, achieving a leading RA score of 1.79. For close-ended questions, Huatuo-34b outperformed GPT-4o with an MCA score of 0.55. In quantitative reasoning, GPT-4o obtained the top RA score of 2.54 and the best MCA score (0.60). Interestingly, a similar trend is witnessed in spatial reasoning, with GPT-4o achieving the best RA score (3.32) in open-ended QA and Claude-Sonnet's 0.48 MCA score in close-ended QA. As for physiological knowledge, Huatuo-34b led with an RA score of 2.08, outperforming other LVLMs.
\begin{figure}[htbp]
\centering
\includegraphics[width=0.5\textwidth]{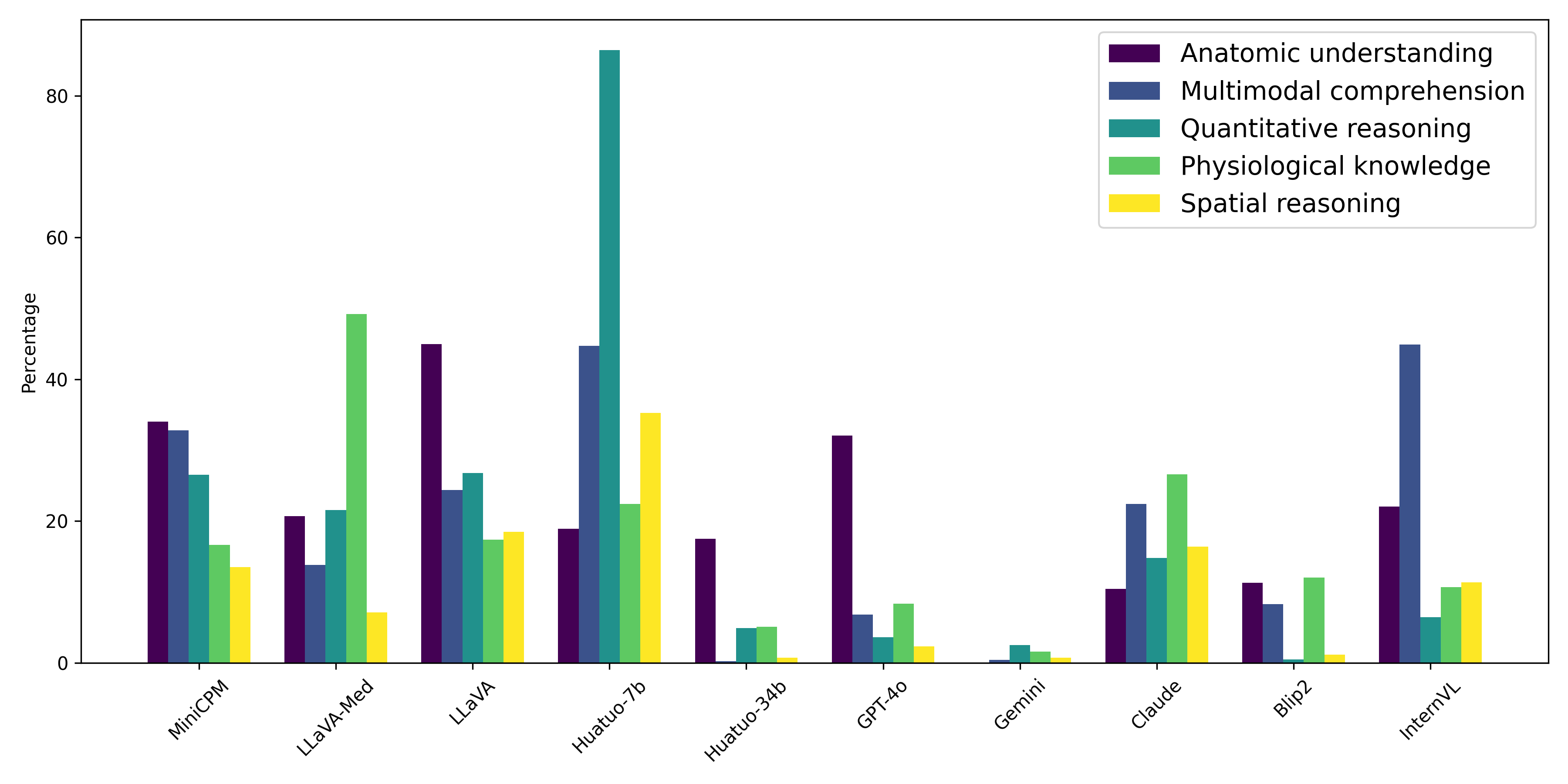}
\caption{Hallucinations of LVLMs on anatomic understanding, multimodal comprehension, quantitative reasoning, physiological knowledge, and spatial reasoning on RadVUQA-CT.}
\label{bar_fig}
\end{figure}

The average hallucination scores were various for all comparisons, ranging from Gemini's 0.7\% to LLaVA's 26.38\% (Table II). To gain a deeper understanding of the hallucination across various clinical aspects, we delve further into the details presented in Fig. \ref{bar_fig}. In particular, half of the LVLMs have the most intensive hallucination in anatomic understanding (MiniCPM, LLaVA, Huatuo-34b, GPT-4o). Huatuo-7b presented the most severe hallucination (86.40\%) in quantitative reasoning (Fig. 3), while this number was dramatically decreased on its 34b variant (4.9\%). Gemini-flash achieved the lowest hallucination scores (0.7\%), with 0.10\%, 0.4\%, 2.5\%, 1.6\%, and 0.7\% on anatomic understanding, multimodal comprehension, quantitative reasoning, physiological knowledge, and spatial reasoning, respectively. Interestingly, although BLIP2 did not outperform other comparisons, it achieved a competitive average hallucination rate, with only 6.63\% of its answers containing hallucinations.

\noindent\textbf{RadVUQA-MRI.} The performance of LVLMs on MRI data presents a similar trend as that on CT, with commercial models dominating the overall performance. GPT-4o achieved state-of-the-art performance with the best score in most evaluation aspects. For instance, it achieved the best score in both multimodal comprehension (2.04 RA and 0.53 MCA) and quantitative reasoning (3.00 RA and 0.57 MCA) QAs. Interestingly, owing to its superior anatomic recognition capabilities on MRI images, GPT-4o achieves better performance in physiological knowledge than Huatuo-34b.

Regarding hallucination rates, Gemini exhibited more hallucinations in MRI data (10.07\%) compared to CT data (0.7\%), whereas GPT-4o demonstrated a significantly lower rate of 0.32\%. Notably, Huatuo-34b showed better stability than GPT-4o and Gemini, with the 2nd least hallucination scores on both CT and MRI data, indicating its more specific training resources.

\begin{table*}[!ht]
\vspace{-0.3cm}
\centering
\caption{Robustness assessment of LVLMs on RadVUQA-OOD}
\resizebox{\textwidth}{!}{%
\begin{threeparttable}
\centering
\begin{tabular}{lcccccccccccccccccc}
\toprule
Model & \multicolumn{3}{c}{Noisy} & \multicolumn{3}{c}{Contrast} & \multicolumn{3}{c}{Sharpness} & \multicolumn{3}{c}{Motion Blur} & \multicolumn{3}{c}{Low Resolution} &
\multicolumn{3}{c}{Histo Equalization} \\
\cmidrule(lr){2-4} \cmidrule(lr){5-7} \cmidrule(lr){8-10} \cmidrule(lr){11-13} \cmidrule(lr){14-16} \cmidrule(lr){17-19}
& OEQ & CEQ & Hallucination & OEQ & CEQ & Hallucination & OEQ & CEQ & Hallucination & OEQ & CEQ & Hallucination & OEQ & CEQ & Hallucination & OEQ & CEQ
& Hallucination\\
\midrule
LLaVA-Med &  1.52 & 0.302 & 19.34\% & 1.59 & 0.332 & 20.99\% & 1.62 & 0.343 & 19.34\% & 1.6 & 0.282 & 22.10\% & 1.64 & 0.362 & 22.10\% & 1.63 & 0.621 & 14.43\%\\
Huatuo-7b & 2.59 & 0.435 & 16.02\% & 2.59 & 0.440 & 21.55\% & 2.67 & 0.428 & 18.89\% & 2.76 & 0.432 & 15\% & 2.7 & 0.442 & 17.13\% & 2.52 & 0.682 & 13.3\% \\
Huatuo-34b & 2.58 & 0.377 & 12.15\% & 2.56 & 0.410 & 19.34\% & 2.61 & 0.397 & 13.26\% & 2.67 & 0.373 & 14.92\% & 2.6 & 0.372 & 17.68\% & 2.48 & 0.665 & 11.7\%\\
\\
LLaVA & 2.27 & 0.435 & 57.46\% & 2.36 & 0.422 & 51.38\% & 2.29 & 0.452 & 56.91\% & 2.48 & 0.442 & 48.07\% & 2.36 & 0.440 & 50.83\% & 2.23 & 0.665 & 19.68\%\\
InternVL & 2.83 & 0.430 & 18.99\% & 2.85 & 0.453 & 17.13\% & 2.82 & 0.455 & 20.99\% & 2.84 & 0.490 & 18.23\% & 2.84 & 0.467 & 18.33\% & 2.56 & 0.663 & \textbf{2.13}\%\\
MiniCPM & 2.41 & 0.445 & 17.13\% & 2.6 & 0.432 & 17.22\% & 2.55 & 0.463 & 12.15\% & 2.54 & 0.467 & 25.41\% & 2.45 & 0.438 & 23.20\% & 2.43 & 0.633 & 10.16\%\\
BLIP2 & 1.04 & 0.187 & \textbf{2.76}\% & 1.04 & 0.168 & 8.83\% & 1.03 & 0.207 & \textbf{3.31}\% & 1.44 & 0.157 & 10.50\% & 1.51 & 0.190 & \textbf{4.97}\% & 1.93 & 0.475& 20.86\%\\
\\
GPT-4o & \textbf{2.92} & 0.608 & 10.50\% & \textbf{2.96} & 0.615 & \textbf{6.67}\% & \textbf{2.97} & 0.607 & 5.52\% & \textbf{3.04} & 0.583 & \textbf{5.52}\% & \textbf{3.02} & 0.618 & 7.18\% & \textbf{2.80} & 0.752 & 3.72\%\\
Gemini-Flash & 2.67 & \textbf{0.743} & 27.78\% & 2.76 & \textbf{0.723} & 27.07\% & 2.64 & \textbf{0.738} & 23.33\% & 2.75 & \textbf{0.733} & 18.78\% & 2.82 & \textbf{0.738} & 24.44\% & 2.78 & \textbf{0.784} & 9.63\%\\
Claude & 2.78 & 0.538 & 11.60\% & 2.86 & 0.447 & 12.71\% & 2.8 & 0.500 & 14.92\% & 2.79 & 0.498 & 12.71\% & 2.79 & 0.478 & 8.38\% & 2.62 & 0.709 & 8.51\%\\
\bottomrule
\end{tabular}
\end{threeparttable}
}
\label{tab: sparse assess}
\vspace{-0.3cm}
\end{table*}

\noindent\textbf{RadVUQA-OOD.} Table III presents the capabilities of LVLMs on the unharmonized data. In particular, GPT-4o achieved the highest RA scores in open-ended questions under all six scenarios, with 2.92, 2.96, 2.97, 3.04, 3.02, and 2.8 on noisy, contrast, sharpness, motion blur, low-resolution, and histogram equalization data respectively). Meanwhile, Gemini-Flash obtained the best MCA scores in all close-ended questions (with 0.743, 0.723, 0.738, 0.733, 0.738, and 0.784, respectively). In addition to GPT-4o and Gemini, InternVL also achieved competitive performance with around 2.83 RA scores. Notably, GPT-4o and BLIP2 were observed to have the lowest hallucination scores for the unharmonized data.

\section{Discussion}

\noindent \textbf{Overall Performance.} The overall performance of LVLMs remains stable on both CT and MRI datasets, with GPT, Gemini, Huatuo (7b and 34b), and Claude achieving the top-5 performance of all comparisons. Specifically, GPT-4o achieved the highest capabilities on both CT and MRI datasets. Huatuo-34 also shows reliable performance in OEQ, making it a strong competitor across different modalities. While models like Gemini and Claude show more variability in their rankings, their presence in the top five for both datasets suggests their high effectiveness. {In particular, the most powerful LVLM (GPT-4o and Gemini) could only achieve comparable results in anatomic understanding and quantitative reasoning, while it underperformed in interpreting linguistic and visual instructions (multimodal comprehension). Interestingly, all models struggled with physiological knowledge, likely due to their limited ability to comprehend visual instructions. Meanwhile, the results indicate that small models (e.g., 7b scales) are prone to hallucination issues. These findings highlight a significant gap in the applicability of current LVLMs to clinical practice, particularly in scenarios requiring the integration of visual and linguistic instructions.

Regarding the trade-offs between model scale and domain-specific knowledge, our findings suggest that, while large closed-source models (e.g., GPT-4) benefit from vast parameter counts and broad general-domain exposure, domain-specific adaptations can help smaller or open-source models (like Huatuo-34b) close the performance gap. However, this relies on the quality and quantity of domain-specific data and may not apply in all cases. For instance, the comparison between LLaVA and its medical adaptation LLaVA -Med shows that even a model originally trained on natural data can achieve better results compared with the one receiving specialized medical data and instruction tuning. This observation indicates that domain-relevant data plays a crucial role—potentially as important as (or even more critical than) raw model scale—in elevating performance on medical imaging tasks. 
The result also suggests that commercial LVLMs continue to outperform their open-source counterparts. Although LLaVA-Med is purportedly better suited for the medical domain, its performance still falls short of generalized open-source LVLMs. Notably, Huatuo-34b demonstrates comparable performance, approaching that of GPT-4o on CT data, despite its smaller model scale. This highlights the potential for medical-specific LVLMs to surpass commercial models as they scale up.

\smallskip
\noindent \textbf{Is LVLM sensitive to CT windowing?} The performance of LVLMs on CT data was widely assessed in existing benchmark studies. Unfortunately, none of them considers the windowing settings while barely collecting 2D CT images (those have been preprocessed) or simply applying min-max normalization. CT windowing is a critical step in clinical practice for optimizing the display of specific tissues. For instance, lung windows highlight pulmonary parenchyma details, while mediastinal windows emphasize soft-tissue contrast. By adjusting the windowing parameters (window level \textbf{WL} and window width \textbf{WW}), radiologists can better differentiate between normal and abnormal tissues, resulting in better diagnosis/screening efficiency. 

\begin{table*}[!htbp]
\caption{Windowing effects on LVLMs}
\centering
\begin{threeparttable}
\begin{tabular}{lcccccc}
\toprule
Models & \makecell{Anatomic \\ understanding} & \makecell{Multimodal\\comprehension} & \makecell{Quantitative reasoning} & \makecell{Spatial reasoning\\} & \makecell{Physiological \\ Knowledge} & \makecell{Average \\Hallucination} \\
\midrule
LLaVA-Med & -0.05 (-5.52\%$^\dagger$) & 0 (-1.35\%) & -1.01$^\dagger$ (-5.41\%$^\dagger$) & 0.14$^\dagger$ (-3.9\%$^\dagger$) & -0.01 (-0.7\%) & -3.38\% \\
Huatuo-7b & 0.08$^\dagger$ (1.69\%) & 0.02 (-0.01\%) & 0.01 (-0.25\%) & -0.03 (-9.98\%$^\dagger$) & 0.02 (-1.5\%) &  -2.01\%\\
Huatuo-34b & 0.08$^\dagger$ (2.29\%$^\dagger$) & -0.01 (0.5\%) & -0.01 (0.25\%) & -0.03 (-0.55\%) & 0 (-0.8\%) & 0.34\%\\
\\
LLaVA & 0.02 (1.46\%) & 0 (2.58\%$^\dagger$) & 0.01 (-0.51\%) & 0.33$^\dagger$ (-12.47\%$^\dagger$) & -0.01 (0.12\%) & -1.76\%\\
InternVL & 0.13$^\dagger$ (-4.26\%$^\dagger$) & 0.06 (0.6\%) & -0.01 (0.3\%) & -0.06 (-2.35\%$^\dagger$) & 0.03 (-0.25\%) & -1.19\%\\
MiniCPM & 0.07$^\dagger$ (-13.56\%$^\dagger$) & -0.02 (0.05\%) & -0.004 (0.08\%) & 0.21$^\dagger$ (-0.56\%) & 0.001 (-0.40) & 2.88\% \\
BLIP2 & -0.13$^\dagger$ (-0.81\%) & 0.1$^\dagger$ (0.9\%) & 0.03 (-0.05\%) & 0.02 (-0.7\%) & -0.49$^\dagger$ (-6.55\%$^\dagger$) & -1.44\%\\
\\
GPT-4o & 0.31$^\dagger$ (-25.4\%$^\dagger$) & 0.01 (-1.8\%) & 0 (-1.0\%) & -0.03 (-1.80\%) & 0 (0.89\%) & -5.82\%\\
Gemini-Flash &  -0.06 (-0.1\%) & -0.03 (-0.4\%)  & 0.04 (-1.2\%) &  0.12$^\dagger$ (-0.7\%) & 0.02 (0\%) & -0.48\%\\
Claude-Sonnet & 0.14$^\dagger$ (-1.6\%) & -0.02(0.2\%) &  0.04 (-3.2\%$^\dagger$) & 0.12$^\dagger$ (-0.16\%) & 0.05 (0.6\%) & -0.83\% \\
\bottomrule
\end{tabular}
\begin{tablenotes}[flushleft]
    \footnotesize
    \item $^\dagger$ refers to significant changes (p\textless0.05) of the results after and before CT windowing. The scores are presented as the residual of open-ended (hallucination) scores between results with CT windowing and w/o CT windowing.
\end{tablenotes}
\end{threeparttable}
\end{table*}

To gain a deeper understanding of how window settings influence LVLMs, we conducted comparative studies on two CT datasets. The control group was derived from 3D CT scans that underwent min-max normalization (referred to as CT-no-winAD), while the comparison group was derived from volumetric CT scans that first underwent window shifting and then min-max normalization. The windowing parameters were set as follows, head and neck(WW:80, WL:40), chest (WW: 1500, WL: -600), abdomen (WW: 400, WL: 50), pelvis, spine and muscles(WW: 400, WL: 40).

Table IV presents the windowing effects on LVLMs, with the values representing the residual in open-ended(hallucination) scores between results with CT windowing and those without CT windowing. Interestingly, the trends between w/o windowing and with windowing were almost consistent. For instance, applying windowing significantly (p\textless0.05) helped LVLMs to improve anatomic understanding (Huatuo-7b, Huatuo-34b, InternVL, MiniCPM, GPT-4o, Claude) and spatial reasoning (LLaVA-Med, LLaVA, MiniCPM, Gemini, Claude), while it might have opposite effects for certain models. In particular, the anatomic understanding and physiological knowledge scores of BLIP2 were notably decreased when applying CT windowing, with -0.13 and -0.49 RA scores, respectively. This suggests that BLIP2 may be trained on pure natural data without any medical samples. 

In addition to RA scores, we found that CT windowing could also alleviate the hallucination issues for most LVLMs. Specifically, the hallucinations of anatomic understanding QA were greatly decreased for LLaVA-Med (5.52\%), MiniCPM (13.56\%), and GPT-4o (25.4\%). In addition to the positive trends, there witnessed few negative effects for Huatuo-34b and LLaVA, with 2.29\% and 2.58\% increments of anatomic understanding and multimodal comprehension, respectively.

\smallskip
\noindent \textbf{Is CoT effective in medical Q\&A?} CoT is a common way to improve the reasoning capabilities of LLM. It involves breaking down complex questions into a series of intermediate steps or smaller sub-tasks, allowing the model to process each step sequentially and produce more accurate and coherent results. To explore whether CoT benefits LVLMs in medical VQA, we construct QA pairs by combining questions with basic or advanced prompts, respectively (basic prompt + OEQs vs advanced prompt + OEQs). 
\begin{table}[!htbp]
\vspace{-0.3cm}
\centering
\caption{Chain of thought assessment on RadVUQA-CT}
\begin{threeparttable}
\centering
\begin{tabular}{lcccc}
\hline
Models & OEQ & OEQ-CoT & CEQ & CEQ-CoT \\
\hline
LLaVA-Med & 1.63 (22.47\%) & 2.08$^\dagger$ (15.76\%$^\dagger$) & 0.29 & 0.39$^\dagger$ \\
Huatuo-7b &  2.15 (41.52\%) & 2.22 (29.42\%$^\dagger$) & 0.44 & 0.48 \\
Huatuo-34b & 2.25 (5.68\%) & 2.33 (4.59\%$^\dagger$) & 0.46 & 0.48 \\
\\
LLaVA & 1.31 (26.38\%) & 1.82$^\dagger$ (19.11\%$^\dagger$) & 0.32 & 0.43$^\dagger$ \\
InternVL &  2.05 (19.07\%)  & 2.18$^\dagger$ (22.32\%$^\dagger$) &  0.49 & 0.48  \\
MiniCPM &  1.86 (24.70\%)  & 2.03$^\dagger$ (23.91\%) & 0.41 & 0.45$^\dagger$ \\
BLIP2 & 1.48 (6.63\%) & 1.72$^\dagger$ (4.98\%$^\dagger$) & 0.29 & 0.34$^\dagger$ \\
\\
GPT-4o &  \textbf{2.49 }(10.61\%) & \textbf{2.64}$^\dagger$ (10.43\%) & \textbf{0.50} & \textbf{0.53}$^\dagger$ \\
Gemini-Flash & 2.34 (\textbf{1.06}\%) & 2.50$^\dagger$ (\textbf{1.47}\%) & 0.46 & 0.52$^\dagger$   \\
Claude-Sonnet & 2.23 (18.12\%) & 2.44$^\dagger$ (22.53\%$^\dagger$)  & 0.45 & 0.52$^\dagger$ \\
\hline
\end{tabular}
\begin{tablenotes}[flushleft]
\footnotesize
\item $^\dagger$ indicates significant changes compared with the one without CoT. The scores are presented as response accuracy (hallucination). 
\end{tablenotes}
\end{threeparttable}
\label{cot}
\vspace{-0.3cm}
\end{table}

As Table \ref{cot} shows, most LVLMs (LLaVA-Med, LLaVA, MiniCPM, BLIP2, and GPT-4o) benefited from prompt CoT in open-ended questions, with higher RA scores (around 0.2 average gain) and lower hallucination frequencies. On the contrary, more hallucinations were observed in certain LVLMs (with 4.4\% of Claude-Sonnet and 0.41\% of Gemini-Flash), indicating that CoT may also mislead LVLMs to misjudgments. Our findings indicate that Chain-of-Thought (CoT) prompting has a varied impact on different reasoning tasks, with significant improvements in certain areas while showing limited effects in others (details in the supplementary materials). Specifically, multimodal comprehension exhibited the most notable enhancement across all LVLMs, with an average increase of 0.40 in RA scores and a 5.59\% reduction in hallucinations. Similarly, spatial reasoning and quantitative reasoning benefited from CoT prompting, with observed RA score gains of 0.28 and 0.29, respectively, and corresponding reductions in hallucinations by 5.1\% and 4.29\%. Interestingly, however, CoT did not lead to significant improvements in the models' physiological knowledge, and the results for anatomic understanding were inconsistent across different models. One possibility is that CoT can occasionally mislead models by reinforcing incorrect intermediate reasoning steps, particularly when the underlying knowledge is uncertain or complex.
For closed-ended questions, in addition to the InternVL and Huatuo variants, the MCA scores of all other LVLMs significantly (p\textless0.05) improved after introducing the chain of thought prompts.

While CoT can enhance reasoning capabilities in certain contexts, its effectiveness appears to vary widely, mainly due to the following reasons 1) The model scale limits the ability to fully exploit CoT reasoning. Small models face parameter capacity limitations, struggling to maintain long-range dependencies and intermediate reasoning steps due to restricted working memory and shallow attention layers.  In contrast, larger, often closed-source models (e.g., 100B+ parameters) tend to benefit more substantially from CoT \cite{singhal2023large}. 2) For essential medical questions (e.g., basic factual queries), applying a chain of thought may not yield significant improvement for some approaches (e.g., Huatuo) since the model’s performance partially depends on knowledge gained during pretraining, rather than elaborate reasoning steps.

These findings highlight the need for a more adaptive CoT approach that tailors prompting strategies based on the complexity and nature of the reasoning task. For instance, leveraging dynamic prompt tuning—where CoT depth and structure are adjusted based on the model’s initial response confidence—could help mitigate potential pitfalls. Future work should explore strategies for optimizing CoT prompting by systematically analyzing failure cases and refining prompt structures to better align with different reasoning modalities. Understanding when and why CoT is beneficial—or detrimental—will be key to maximizing its potential across diverse clinical and multimodal AI applications.

\smallskip
\noindent \textbf{Are LVLMs robust to unharmonized data?} 
Table III demonstrates the performance of LVLMs on unharmonized data. To further explore the effect of different variants, we compared the LVLMs' performance (on unharmonized data, e.g., noise, blur, and contrast adjustments) against those on the raw data. Interestingly, we found that the tolerances of different LVLMs for unharmonized data were various. For instance, while Huatuo (both 7B and 34B) demonstrated reduced accuracy on augmented data, it experienced a decline in the rate of hallucinations. On the contrary, although Gemini achieved slightly higher accuracy scores in close-ended questions, it dramatically encountered hallucination issues (from an average hallucination of 0.7\% to 24.28\%). Notably, InternVL demonstrated the most stable results, with the least performance shift observed on the unharmonized data, indicating a strong robustness to data variation. 

To further explore data harmonization effects, we applied histogram equalization as a preprocessing step (shown in Table III). Interestingly, most LVLMs showed considerable improvements in closed-ended question (CEQ) performance under this setting. Notably, Gemini-Flash reached a 0.784 MCA score—an increase of approximately 10\%—making it the top-performing model for CEQs. However, the same preprocessing appeared to hinder open-ended question performance for all but BLIP2. This discrepancy may arise from BLIP2’s training procedure, which potentially included histogram-equalized images, thus enabling the model to adapt more readily. Overall, these results highlight that while histogram equalization can enhance certain tasks, its efficacy may vary depending on the model’s training protocol and the nature of the queries.

This suggests that future research must address these gaps by systematically considering input data characteristics, therefore harmonizing the input data to the same data space. This involves developing harmonization foundation models that resolve the heterogeneity of medical imaging practices for multi-center datasets. By doing so, LVLMs can not only handle in-domain queries but also achieve reliable and effective results in diverse clinical settings.

\smallskip
\noindent \textbf{Can LVLMs distinguish synthetic data from the real ones?} To investigate whether LVLMs could distinguish synthesized data, we calculated the synthetic-detection rates of LVLMs. Our finding (Table VI) suggests that LVLMs struggle with synthetic data, even when the synthetic one does not follow the human physiological structures (Fig. \ref{fig4}). This is particularly true when the synthetic data is generated using advanced techniques that replicate the intricate patterns and textures of real data. 

Specifically, all LVLMs exhibited low synthetic-detection rates, ranging from 0\% to 35.2\%. However, it is worth noting that some models adopt a more sophisticated approach to handle this uncertainty. For example, GPT-4o stands out by refusing to answer nearly all queries related to synthetic data identification. This behavior suggests a better safety mechanism, where the model opts for caution rather than providing potentially inaccurate information. Such a refusal to respond may be interpreted as acknowledging the model's limitations, allowing it to avoid the risk of hallucination or misclassification. Moreover, all LVLMs showed weak capabilities in distinguishing nonhuman data, with RA scores of 1.03 (BLIP2) to 2.27 (LLaVA-Med).

Our findings suggest that current LVLMs struggle to reliably distinguish synthetic images, raising important questions about how to maintain data integrity in a clinical setting. The detection of synthetic data may be addressed in the following ways: 1) Assessing data distribution.
Normally, synthetic data lacks the natural variability and noise present in real-world data, this includes: uniform distributions where real data would have peaks and valleys, unusual consistency in patterns, a lack of outliers, and easily identifiable repeating sequences; 2) Watermark detection. Using invisible, machine-readable watermarks in synthetic data can provide a clear signal for downstream models or validation tools. When a watermark is detected, the model or system could raise a flag or automatically classify the input as synthetic. However, this requires a recognized standard protocol for all generative models. 3) Fine-tuning with labeled synthetic Data. Including explicitly labeled synthetic and real examples during training (or fine-tuning) may help models learn the nuanced distinctions. However, this is inapplicable as it largely increases the cost and it is almost impossible to collect all generative models for training.

\begin{figure}[htbp]
\centering
\includegraphics[width=0.4\textwidth]{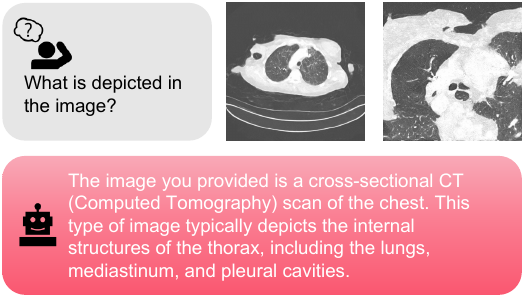}
\caption{Weak capability of LVLMs when testing on low-quality synthetic data.}
\label{fig4}
\end{figure}

\begin{table}[!htbp]
\vspace{-0.3cm}
\centering
\caption{Synthetic and nonhuman data assessment}
\begin{threeparttable}
\centering
\setlength{\tabcolsep}{14pt}
\begin{tabular}{lcc}
\toprule
Model & Synthetic data & NonHuman data\\
\midrule
LLaVA-Med  & 7.6\% [43.6\%] & 2.27 (16.13\%) \\
Huatuo-7b      & 14.8\% [0.4\%]  & 1.82 (30.64\%) \\
Huatuo-34   & 35.2\% [49.2\%] & 1.85 (12.90\%) \\
\\
LLaVA     & 8.5\% [0\%]     & 1.47 (24.19\%) \\
InternVL    & 6\% [0.8\%]   & 1.49 (36.07\%) \\
MiniCPM     & 7.6\% [43.6\%] & 1.53 (19.35\%) \\
BLIP2        & 0 [26.8\%]     & 1.03 (6.45\%) \\
\\
GPT-4o         & 0 [99.2\%]     & 1.74 (19.35\%) \\
Gemini-Flash      & 0 [0\%]      & 2.01 (6.45\%) \\
\bottomrule
\end{tabular}
\begin{tablenotes}[flushleft]
\footnotesize
\item The results are shown in synthetic-detection rate [rejection rate] for synthetic data, and RA score (hallucination score). 
\end{tablenotes}
\end{threeparttable}
\label{oodtestresults}
\vspace{-0.3cm}
\end{table}

\smallskip
\noindent \textbf{Severe hallucinations render LVLMs inapplicable.} Hallucination issue is particularly concerning in the medical domain, where accuracy and reliability are paramount. It can lead to misleading conclusions, erroneous diagnoses, or inappropriate treatment recommendations when LVLMs are used to analyze medical data or assist in decision-making processes. 

By leveraging refusal as a strategy, models like GPT-4o may demonstrate an advanced level of decision-making that prioritizes model safety over attempting to provide an answer at all costs. This could be particularly useful in high-stakes applications, where the cost of an incorrect prediction is significant. Therefore, while most models still struggle with distinguishing synthetic data, integrating a refusal-to-answer strategy may represent a step towards building more robust and safer AI systems. Furthermore, this could be also achieved by incorporating invisible watermarks in generative models or integrating additional detection algorithms.

\noindent \textbf{Ethical issues of LVLMs.} As existing LVLMs raise concerns about patient safety, liability, and public trust in AI-assisted medicine, we propose possible safeguards and regulatory considerations.
Regulatory Oversight and Validation. Clinical Trials and Approval Processes: Before integrating LVLMs into clinical workflows, conducting rigorous validation studies—akin to clinical trials—can help quantify safety, efficacy, and error rates. 
Fail-Safe Mechanisms and Refusal to Answer. Confidence Estimation: Beyond the refusal-to-answer strategy seen in chatgpt, providing models with a mechanism to estimate their uncertainty can help them identify situations where they lack sufficient knowledge or clarity. This can be achieved through probabilistic outputs, confidence scores, or specialized OOD (Out-of-Distribution) detectors that signal when inputs deviate from the training domain.
Human-in-the-Loop Verification: When the model’s confidence falls below a certain threshold, results should be flagged for clinician review. 
Ethical and Policy Recommendations. The consortium should establish accountability and liability for LVLMs. Clear guidelines are needed to define who is responsible if an LVLM’s recommendation leads to adverse patient outcomes. 

\noindent \textbf{Future perspectives and research direction}. In addition to the current support for MRI data, our future endeavors aim to extend RadVUQA to include more imaging modalities—particularly ultrasound and X-ray. By incorporating these modalities, we hope to capture a broader spectrum of clinical scenarios and address modality-specific nuances, ultimately enhancing the generalizability and robustness of our benchmark. We plan to establish high-quality, annotated datasets for both ultrasound and X-ray, refining our evaluation methodologies to accommodate each modality’s distinct imaging characteristics. Additionally, longitudinal studies may be integrated into future iterations of RadVUQA, offering insights into disease progression and treatment responses over time.
Beyond expanding modality coverage, interdisciplinary collaborations between AI experts and medical professionals will be crucial for developing clinically viable LVLMs. Close partnerships can help define clinically relevant tasks, annotate complex cases, and align model outputs with routine diagnostic workflows. Such synergy ensures that benchmarks like RadVUQA remain meaningful to end users, rather than purely academic exercises.

Meanwhile, emerging technologies such as explainable AI (XAI) \cite{xai-cortinas2023toward, tjoa2020survey} and federated learning \cite{chen2023metafed} hold significant promise for advancing LVLMs in real-world environments. Explainable AI offers pathways to interpret and visualize model decisions, which is essential for earning clinician trust and meeting regulatory expectations regarding transparency and accountability. Federated learning, on the other hand, addresses data privacy concerns by training models on distributed healthcare data without transferring sensitive patient information to a central server. This could both increase the diversity and volume of training data (boosting model robustness) while adhering to strict privacy regulations.

\section{Conclusion}
This paper presented a novel benchmark study, \textbf{Rad}iological \textbf{V}ision \textbf{U}nderstanding, \textbf{Q}uestioning and \textbf{A}nswering, for evaluating large vision language models through in-depth clinical-specific perspectives. Based on our comprehensive assessments, we claim that 
\begin{itemize}
    \item Large-scale medical LVLMs remain underdeveloped but hold significant potential to outperform commercial models.
    \item Clinical-specific techniques such as CT windowing could improve LVLMs' capabilities and alleviate hallucination issues.
    \item Prompt-CoT is effective in enhancing model capabilities with fewer hallucinations and higher accuracy.
    \item LVLMs are not robust to unharmonized data, even certain models have lower hallucinations on the low-quality data.
    \item LVLMs are struggling to distinguish synthetic or nonhuman data, thus adversarial attacks could bring serious safety challenges. 
\end{itemize}
A startling finding is that even though some LVLMs claim to be capable of diagnostics and report generation, they lack a basic understanding of anatomical structures. This reveals that current LVLMs are neither generalizable nor ready for clinical application. Moreover, it highlights a critical gap in their foundational medical knowledge, underscoring the need for the development of more versatile and robust models. We believe our findings offer important insights that will help guide future research in advancing LVLMs for medical use.

\bibliographystyle{IEEEtran}
\bibliography{ref}

\begin{thebibliography}{10}
\providecommand{\url}[1]{#1}
\csname url@samestyle\endcsname
\providecommand{\newblock}{\relax}
\providecommand{\bibinfo}[2]{#2}
\providecommand{\BIBentrySTDinterwordspacing}{\spaceskip=0pt\relax}
\providecommand{\BIBentryALTinterwordstretchfactor}{4}
\providecommand{\BIBentryALTinterwordspacing}{\spaceskip=\fontdimen2\font plus
\BIBentryALTinterwordstretchfactor\fontdimen3\font minus \fontdimen4\font\relax}
\providecommand{\BIBforeignlanguage}[2]{{%
\expandafter\ifx\csname l@#1\endcsname\relax
\typeout{** WARNING: IEEEtran.bst: No hyphenation pattern has been}%
\typeout{** loaded for the language `#1'. Using the pattern for}%
\typeout{** the default language instead.}%
\else
\language=\csname l@#1\endcsname
\fi
#2}}
\providecommand{\BIBdecl}{\relax}
\BIBdecl

\bibitem{vqarad}
J.~J. Lau, S.~Gayen, A.~Ben~Abacha, and D.~Demner-Fushman, ``A dataset of clinically generated visual questions and answers about radiology images,'' \emph{Scientific data}, vol.~5, no.~1, pp. 1--10, 2018.

\bibitem{slake}
B.~Liu, L.-M. Zhan, L.~Xu, L.~Ma, Y.~Yang, and X.-M. Wu, ``Slake: A semantically-labeled knowledge-enhanced dataset for medical visual question answering,'' in \emph{2021 IEEE 18th International Symposium on Biomedical Imaging (ISBI)}.\hskip 1em plus 0.5em minus 0.4em\relax IEEE, 2021, pp. 1650--1654.

\bibitem{omnimedvqa}
Y.~Hu, T.~Li, Q.~Lu, W.~Shao, J.~He, Y.~Qiao, and P.~Luo, ``Omnimedvqa: A new large-scale comprehensive evaluation benchmark for medical lvlm,'' in \emph{Proceedings of the IEEE/CVF Conference on Computer Vision and Pattern Recognition}, 2024, pp. 22\,170--22\,183.

\bibitem{clip}
A.~Radford, J.~W. Kim, C.~Hallacy, A.~Ramesh, G.~Goh, S.~Agarwal, G.~Sastry, A.~Askell, P.~Mishkin, J.~Clark \emph{et~al.}, ``Learning transferable visual models from natural language supervision,'' in \emph{International conference on machine learning}.\hskip 1em plus 0.5em minus 0.4em\relax PMLR, 2021, pp. 8748--8763.

\bibitem{alayrac2022flamingo}
J.-B. Alayrac, J.~Donahue, P.~Luc, A.~Miech, I.~Barr, Y.~Hasson, K.~Lenc, A.~Mensch, K.~Millican, M.~Reynolds \emph{et~al.}, ``Flamingo: a visual language model for few-shot learning,'' \emph{Advances in neural information processing systems}, vol.~35, pp. 23\,716--23\,736, 2022.

\bibitem{li2023blip}
J.~Li, D.~Li, S.~Savarese, and S.~Hoi, ``Blip-2: Bootstrapping language-image pre-training with frozen image encoders and large language models,'' in \emph{International conference on machine learning}.\hskip 1em plus 0.5em minus 0.4em\relax PMLR, 2023, pp. 19\,730--19\,742.

\bibitem{llava}
H.~Liu, C.~Li, Q.~Wu, and Y.~J. Lee, ``Visual instruction tuning,'' \emph{Advances in neural information processing systems}, vol.~36, 2024.

\bibitem{internvl}
Z.~Chen, J.~Wu, W.~Wang, W.~Su, G.~Chen, S.~Xing, M.~Zhong, Q.~Zhang, X.~Zhu, L.~Lu \emph{et~al.}, ``Internvl: Scaling up vision foundation models and aligning for generic visual-linguistic tasks,'' in \emph{Proceedings of the IEEE/CVF Conference on Computer Vision and Pattern Recognition}, 2024, pp. 24\,185--24\,198.

\bibitem{medflamingo}
M.~Moor, Q.~Huang, S.~Wu, M.~Yasunaga, Y.~Dalmia, J.~Leskovec, C.~Zakka, E.~P. Reis, and P.~Rajpurkar, ``Med-flamingo: a multimodal medical few-shot learner,'' in \emph{Machine Learning for Health (ML4H)}.\hskip 1em plus 0.5em minus 0.4em\relax PMLR, 2023, pp. 353--367.

\bibitem{medllava}
C.~Li, C.~Wong, S.~Zhang, N.~Usuyama, H.~Liu, J.~Yang, T.~Naumann, H.~Poon, and J.~Gao, ``Llava-med: Training a large language-and-vision assistant for biomedicine in one day,'' \emph{Advances in Neural Information Processing Systems}, vol.~36, 2024.

\bibitem{PMC}
S.~Zhang, Y.~Xu, N.~Usuyama, J.~Bagga, R.~Tinn, S.~Preston, R.~Rao, M.~Wei, N.~Valluri, C.~Wong \emph{et~al.}, ``Large-scale domain-specific pretraining for biomedical vision-language processing,'' \emph{arXiv preprint arXiv:2303.00915}, vol.~2, no.~3, p.~6, 2023.

\bibitem{radfm}
C.~Wu, X.~Zhang, Y.~Zhang, Y.~Wang, and W.~Xie, ``Towards generalist foundation model for radiology,'' \emph{arXiv preprint arXiv:2308.02463}, 2023.

\bibitem{chen2024huatuogpt}
J.~Chen, R.~Ouyang, A.~Gao, S.~Chen, G.~H. Chen, X.~Wang, R.~Zhang, Z.~Cai, K.~Ji, G.~Yu \emph{et~al.}, ``Huatuogpt-vision, towards injecting medical visual knowledge into multimodal llms at scale,'' \emph{arXiv preprint arXiv:2406.19280}, 2024.

\bibitem{bai2023qwen}
J.~Bai, S.~Bai, Y.~Chu, Z.~Cui, K.~Dang, X.~Deng, Y.~Fan, W.~Ge, Y.~Han, F.~Huang \emph{et~al.}, ``Qwen technical report,'' \emph{arXiv preprint arXiv:2309.16609}, 2023.

\bibitem{young2024yi}
A.~Young, B.~Chen, C.~Li, C.~Huang, G.~Zhang, G.~Zhang, H.~Li, J.~Zhu, J.~Chen, J.~Chang \emph{et~al.}, ``Yi: Open foundation models by 01. ai,'' \emph{arXiv preprint arXiv:2403.04652}, 2024.

\bibitem{he2020pathvqa}
X.~He, Y.~Zhang, L.~Mou, E.~Xing, and P.~Xie, ``Pathvqa: 30000+ questions for medical visual question answering,'' \emph{arXiv preprint arXiv:2003.10286}, 2020.

\bibitem{xia2024cares}
P.~Xia, Z.~Chen, J.~Tian, Y.~Gong, R.~Hou, Y.~Xu, Z.~Wu, Z.~Fan, Y.~Zhou, K.~Zhu \emph{et~al.}, ``Cares: A comprehensive benchmark of trustworthiness in medical vision language models,'' \emph{arXiv preprint arXiv:2406.06007}, 2024.

\bibitem{wasserthal2023totalsegmentator}
J.~Wasserthal, H.-C. Breit, M.~T. Meyer, M.~Pradella, D.~Hinck, A.~W. Sauter, T.~Heye, D.~T. Boll, J.~Cyriac, S.~Yang \emph{et~al.}, ``Totalsegmentator: robust segmentation of 104 anatomic structures in ct images,'' \emph{Radiology: Artificial Intelligence}, vol.~5, no.~5, 2023.

\bibitem{d2024totalsegmentator}
T.~A. D'Antonoli, L.~K. Berger, A.~K. Indrakanti, N.~Vishwanathan, J.~Wei{\ss}, M.~Jung, Z.~Berkarda, A.~Rau, M.~Reisert, T.~K{\"u}stner \emph{et~al.}, ``Totalsegmentator mri: Sequence-independent segmentation of 59 anatomical structures in mr images,'' \emph{arXiv preprint arXiv:2405.19492}, 2024.

\bibitem{xing2023less}
X.~Xing, G.~Papanastasiou, S.~Walsh, and G.~Yang, ``Less is more: unsupervised mask-guided annotated ct image synthesis with minimum manual segmentations,'' \emph{IEEE Transactions on Medical Imaging}, vol.~42, no.~9, pp. 2566--2576, 2023.

\bibitem{soares2020sars}
E.~Soares, P.~Angelov, S.~Biaso, M.~H. Froes, and D.~K. Abe, ``Sars-cov-2 ct-scan dataset: A large dataset of real patients ct scans for sars-cov-2 identification,'' \emph{MedRxiv}, pp. 2020--04, 2020.

\bibitem{nan2022data}
Y.~Nan, J.~Del~Ser, S.~Walsh, C.~Sch{\"o}nlieb, M.~Roberts, I.~Selby, K.~Howard, J.~Owen, J.~Neville, J.~Guiot \emph{et~al.}, ``Data harmonisation for information fusion in digital healthcare: A state-of-the-art systematic review, meta-analysis and future research directions,'' \emph{Information Fusion}, vol.~82, pp. 99--122, 2022.

\bibitem{yao2024minicpm}
Y.~Yao, T.~Yu, A.~Zhang, C.~Wang, J.~Cui, H.~Zhu, T.~Cai, H.~Li, W.~Zhao, Z.~He \emph{et~al.}, ``Minicpm-v: A gpt-4v level mllm on your phone,'' \emph{arXiv preprint arXiv:2408.01800}, 2024.

\bibitem{kwon2023efficient}
W.~Kwon, Z.~Li, S.~Zhuang, Y.~Sheng, L.~Zheng, C.~H. Yu, J.~E. Gonzalez, H.~Zhang, and I.~Stoica, ``Efficient memory management for large language model serving with pagedattention,'' in \emph{Proceedings of the ACM SIGOPS 29th Symposium on Operating Systems Principles}, 2023.

\bibitem{zheng2024judging}
L.~Zheng, W.-L. Chiang, Y.~Sheng, S.~Zhuang, Z.~Wu, Y.~Zhuang, Z.~Lin, Z.~Li, D.~Li, E.~Xing \emph{et~al.}, ``Judging llm-as-a-judge with mt-bench and chatbot arena,'' \emph{Advances in Neural Information Processing Systems}, vol.~36, 2024.

\bibitem{chen2024mllm}
D.~Chen, R.~Chen, S.~Zhang, Y.~Liu, Y.~Wang, H.~Zhou, Q.~Zhang, P.~Zhou, Y.~Wan, and L.~Sun, ``Mllm-as-a-judge: Assessing multimodal llm-as-a-judge with vision-language benchmark,'' \emph{arXiv preprint arXiv:2402.04788}, 2024.

\bibitem{singhal2023large}
K.~Singhal, S.~Azizi, T.~Tu, S.~S. Mahdavi, J.~Wei, H.~W. Chung, N.~Scales, A.~Tanwani, H.~Cole-Lewis, S.~Pfohl \emph{et~al.}, ``Large language models encode clinical knowledge,'' \emph{Nature}, vol. 620, no. 7972, pp. 172--180, 2023.

\bibitem{xai-cortinas2023toward}
K.~Corti{\~n}as-Lorenzo and G.~Lacey, ``Toward explainable affective computing: A review,'' \emph{IEEE Transactions on Neural Networks and Learning Systems}, 2023.

\bibitem{tjoa2020survey}
E.~Tjoa and C.~Guan, ``A survey on explainable artificial intelligence (xai): Toward medical xai,'' \emph{IEEE transactions on neural networks and learning systems}, vol.~32, no.~11, pp. 4793--4813, 2020.

\bibitem{chen2023metafed}
Y.~Chen, W.~Lu, X.~Qin, J.~Wang, and X.~Xie, ``Metafed: Federated learning among federations with cyclic knowledge distillation for personalized healthcare,'' \emph{IEEE Transactions on Neural Networks and Learning Systems}, 2023.

\end{thebibliography}
\end{document}